\documentclass[IJPHM, 2014, 00]{PHMSociety}

\usepackage{graphicx}
\usepackage{amsmath}
\usepackage{float} % For the 'H' specifier

\usepackage{booktabs}
\newcommand{\ra}[1]{\renewcommand{\arraystretch}{#1}}

\begin{document}

% Paper Title
\title{A Generic Machine Learning Framework for Fully-Unsupervised Anomaly Detection with Contaminated Data}

% Authors List
\author{%			
	Markus Ulmer\authorNumber{1}, Jannik Zgraggen\authorNumber{2}, and Lilach Goren Huber\authorNumber{3}
}

% Author Affiliations
\address{% This is a tabular environment so each affiliation needs to be separated by "\\" or "\tabularnewline"
	\affiliation{{1,2,3}}{Zurich University of Applied Sciences, Winterthur 8401, Switzerland}{ %add emails
		{\email{markus.ulmer@zhaw.ch}}\\ 
		{\email{jannik.zgraggen@zhaw.ch}}\\ 
		{\email{lilach.gorenhuber@zhaw.ch}}
		} % emails input

}

% Create the title
\maketitle
\pagestyle{fancy}
\thispagestyle{plain}

% Abstract
\begin{abstract}%   %NOTE: Deleting the percentage after "{abstract}" may be lead to an extra leading space in the first line of the abstract, and this should be prevented.
Anomaly detection (AD) tasks have been solved using machine learning algorithms in various domains and applications. The great majority of these algorithms use normal data to train a residual-based model and assign anomaly scores to unseen samples based on their dissimilarity with the learned normal regime. The underlying assumption of these approaches is that anomaly-free data is available for training. This is, however, often not the case in real-world operational settings, where the training data may be contaminated with an unknown fraction of abnormal samples. Training with contaminated data, in turn, inevitably leads to a deteriorated AD performance of the residual-based algorithms. 

In this paper we introduce a framework for a fully unsupervised refinement of contaminated training data for AD tasks. The framework is generic and can be applied to any residual-based machine learning model. We demonstrate the application of the framework to two public datasets of multivariate time series machine data from different application fields. We show its clear superiority over the naive approach of training with contaminated data without refinement. Moreover, we compare it to the ideal, unrealistic reference in which anomaly-free data would be available for training. The method is based on evaluating the contribution of individual samples to the generalization ability of a given model, and contrasting the contribution of anomalies with the one of normal samples. As a result, the proposed approach is comparable to, and often outperforms training with normal samples only. 
\end{abstract}

%%%%%%%%%%%%%%%%%%%%%%%%%%%%%%%%%%%%%
%%%%%%%%%%%%%%%%%%%%%%%%%%%%%%%%%%%%%%
\section{Introduction}
%%%%%%%%%%%%%%%%%%%%%%%%%%%%%%%%%%%%%
%%%%%%%%%%%%%%%%%%%%%%%%%%%%%%%%%%%%%%

Anomaly detection (AD) tasks are common in very diverse fields, including medical image processing, autonomous driving, fraud detection, and fault detection in industrial machines. 
An inherent property of AD tasks is that very few or no labeled examples of anomalous behavior are provided in advance. 

Therefore, the most common machine learning approaches to solve these tasks are based on using exclusively normal data to train a selected prediction algorithm, which is subsequently used to infer on unseen data. The underlying assumption here is that the algorithm's prediction errors (residuals) will be higher whenever the input sample does not belong to the learned distribution. This family of models can be referred to as residual-based models, irrespective of whether they use regression or reconstruction residuals to detect anomalies, with the most common models being reconstruction models such as principal component analysis (PCA) or various types of autoencoder (AE) neural networks. It is worth noting that these models are often termed "unsupervised". However, in the context of AD, they should be referred to as "semi supervised" since they assume the availability of labeled normal data for training.   

In fact, such models tend to perform rather poorly whenever contamination in the form of anomalous samples is introduced into the training data. In real-world applications, however, the assumption of having anomaly-free training data does not always hold, as data contamination cannot be avoided. 
In this case, truly unsupervised methods, based on clustering or one-class classification \cite{scholkopf1999support}, are required. Recently there has been a growing effort to develop deep unsupervised algorithms for AD, whose performance is not severely damaged by data contamination \cite{munir2018deepant}.

Despite the high practical relevance of the problem, systematic solutions for AD with contaminated training data are rather rare. Recently, several papers have suggested useful approaches based on data refinement \cite{yoon2021self}, on latent outlier exposure \cite{qiu2022latent}, and on physics-informed deep learning \cite{zgraggen2023fully}. 

In this paper, we propose a novel framework that addresses the challenge of AD with potentially contaminated training data. Our approach here is generic: the framework can be used with any residual-based machine learning model (e.g. PCA, any kind of AE or regression NN). Our aim is to offer a substantial improvement of the residual-based approach by allowing any residual-based AD model (intended to be trained on anomaly-free data) to be applicable in a fully unsupervised setting without any labels and not assuming anomaly-free training data.

The framework is based on a single-step (not iterative) refinement of the contaminated data which proposes candidate anomalies to be removed from the training data. The refinement algorithm itself does not depend on the contamination ratio. The only assumption behind it is that the majority class is the normal one (which is a defining property of any AD task).

We demonstrate the performance of the proposed framework on two industrial time series datasets. The first dataset is of high-frequency acoustic sensor data converted to Mel Spectrograms. With this dataset, we address the use-case of abrupt machine faults of various types. The second dataset is a multivariate time series from Turbofan aircraft engines aimed at monitoring the gradual degradation of the engines over their lifetime.  

For each dataset, we evaluate the data refinement quality by comparing to the naive approach of training a residual-based model blindly with the entire contaminated data. For reference, we also compare the performance to the ideal (but rather unrealistic) case of training with anomaly-free data. We show that the refinement step is essential in order to achieve high performance AD in the presence of significant contamination. 

The contribution of the paper is in addressing the highly relevant challenge of AD under realistic conditions in which anomaly-free training data is not available. The approach we take is simple for implementation with any existing residual-based AD model, and can be applied on raw data or on learnable representations. Yet, it is shown to perform as well as the ideal reference of training the same model with anomaly-free data and sometimes outperforms it in its refinement efficacy, as it utilizes the properties of the anomalous samples and contrasts them with the normal samples. The focus of this paper is on the effectiveness of the framework for AD on time series data. However, the framework is generic in nature and should in principle be applicable to any data type. 

%Contributions:

%- Address a practically relevant problem that is rarely solved in the literature.

%- Simple but effective refinement framework.

%- Utilizes properties of the unknown anomalies, and contrasts them directly with those of the normal samples. 

%- does not rely on knowing the contamination ratio. 

%- Model agnostic, classical ML or sophisticated deep networks. 

%- In principle for any data type. 

%- in line with the growing trend of data centric AI (cite Andrew Ng).

%%%%%%%%%%%%%%%%%%%%%%%%%%%%%%%%%%%%%
%%%%%%%%%%%%%%%%%%%%%%%%%%%%%%%%%%%%%%
\section{Related Work}\label{sec:related}
%%%%%%%%%%%%%%%%%%%%%%%%%%%%%%%%%%%%%
%%%%%%%%%%%%%%%%%%%%%%%%%%%%%%%%%%%%%
\paragraph{AD with machine learning.}
A large variety of machine learning methods for AD have been developed in recent years. The great majority of the work has focused on standard AD problems, in which the training data is assumed to be anomaly-free. 
Many of the classical algorithms are distance or density-based, like one-class classifiers \cite{scholkopf1999support,tax2004support} or density estimation methods \cite{latecki2007outlier}. 
Later works suggest various deep learning models, mostly based on autoencoders (AEs) that are  trained on normal data and detect anomalies based on the model residuals at inference. Such models  have been applied to various data types including images \cite{zhou2017anomaly} and multi-variate time series \cite{audibert2020usad,munir2018deepant,zhang2019deep}. 

More recently it has been demonstrated that state-of-the-art AD performance can be achieved by one-class classifiers on pre-trained features extracted from deep learning architectures \cite{sohn2020learning}. Different variants of self-supervised feature extractors have been exploited for AD on image data \cite{golan2018deep,bergman2020classification, hendrycks2018deep} as well as on tabular and time series data \cite{shenkar2022anomaly,schneider2022detecting,michau2022fully}.

As explained above, the common prerequisite to all of the above methods is anomaly-free ("normal") training data. Since this assumption is rarely valid in practical settings, the present paper deals with the challenge of a contaminated training dataset. 

\paragraph{AD with contaminated training data.}
The most common approach to address the problem of anomaly-contamination in the training data is to assume that the fraction of anomalies is low enough, so that the standard AD algorithms can be "blindly" trained on the contaminated data \cite{zong2018deep,bergman2020classification}. However, as has been shown before, this assumption breaks down already for rather low contamination ratios (e.g $5\%$ with robust approaches \cite{munir2018deepant}).

An alternative approach to deal with contaminated data are iterative refinement methods, which use one-class classifiers (OCCs) \cite{beggel2020robust} or reconstruction errors of the AE \cite{zhou2017anomaly,berg2019unsupervised} in order to remove the anomalies and improve the AD performance subsequently. \cite{yoon2021self} suggests to boost the refinement performance yet more using an ensemble of OCCs instead of a single model. 
Recent papers \cite{qiu2022latent,wang2022hierarchical} suggest that contrasting information from the anomalous class helps to improve the AD performance with contaminated training data. 

Our approach makes conceptual use of ideas of the above papers, but suggests a new simple approach, inspired by data centric concepts. The proposed framework refines the contaminated training data by splitting it into partially overlapping subsets and training an ensemble of residual based models. The refinement score assigned to each sample contains information about its contribution to the generalization ability of the trained model. This is done by contrasting the AD performance of ensemble members trained with and without this sample in the training set. Our approach differs from the ones mentioned above by its simplicity and by its generic nature. The idea can be applied with any residual-based model, be it a reconstruction model (PCA or any variant of AE) or a regression model (from linear regression to deep neural networks). The suggested framework has no assumption regarding the contamination rate, and relies solely on the basic anomaly-detection assumption of a normal majority class. Moreover, the method is demonstrated here for time series data, but should be conceptually valid for any data type. In this sense the framework is very compatible with the recent trend of data centric AI approaches.

\begin{figure*}[tb]
\begin{center}
\includegraphics[width=0.9\textwidth]{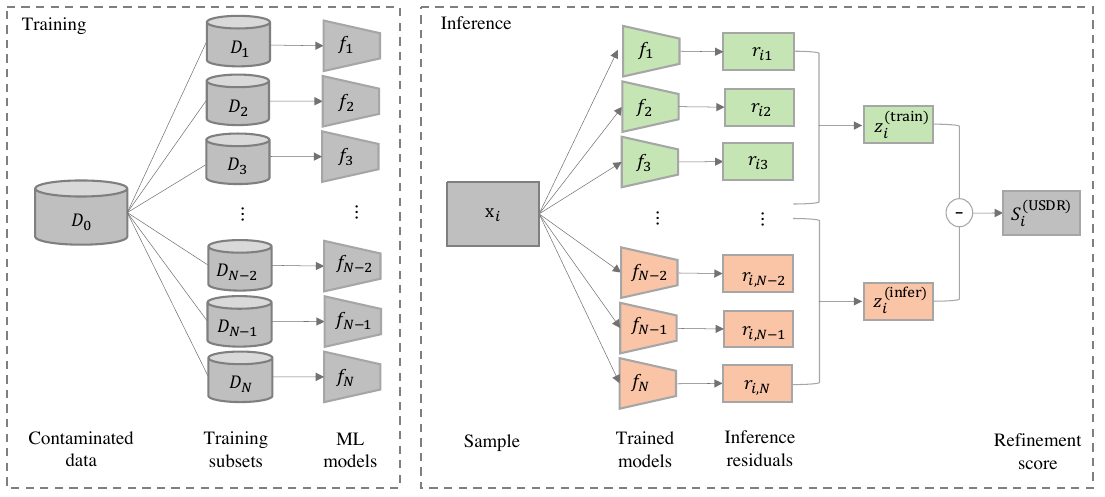}
\caption{The proposed Unsupervised Data Refinement framework.  }
\label{fig:framework}
\end{center}
\end{figure*}
%%%%

%%%%%%%%%%%%%%%%%%%%%%%%%%%%%%%%%%%%%
%%%%%%%%%%%%%%%%%%%%%%%%%%%%%%%%%%%%%%
\section{Method}\label{sec:method}
%%%%%%%%%%%%%%%%%%%%%%%%%%%%%%%%%%%%%
%%%%%%%%%%%%%%%%%%%%%%%%%%%%%%%%%%%%%

We assume a training dataset $\mathcal{D}_0=\{ {\mathbf x}_i,{\mathbf y}_i \}_{i=1}^{N}$ containing $N$ samples of normal data with abnormal contamination where ${\mathbf x}_i$ are the input and ${\mathbf y}_i$ the output variables. In contrast to the common notation in AD problems, here the target variables ${\mathbf y}_i$ are not the outputs of a binary classifier that can only obtain the values $0$ and $1$. The reason is, that our method suggests a generic framework that applies to any residual-based prediction model ${\mathbf y}_i=f({\mathbf x}_i)$ which is traditionally trained with normal data. Note that in case of a reconstruction model (like PCA or AE), the target variables ${\mathbf y}_i$ are equal to ${\mathbf x}_i$ and the predictions are thus $\hat{\mathbf x}_{i}$. On the other hand, in a regression setup, the outputs ${\mathbf y}_i$ are typically different from the inputs ${\mathbf x}_i$. In any case, no normal/abnormal labels are assumed during the entire training and inference process, and in this sense the data is unlabeled. 

\subsection{Proposed Framework}
The goal of the proposed method is to refine the unlabeled contaminated training data $\mathcal{D}_0$, such that at a second step only normal samples can be used to ideally train a residual-based model of choice. 

The suggested USDR framework is model agnostic. Any residual-based model can be selected, and is trained with multiple subsets of the original training data $\mathcal{D}_0$. When inferring with the resulting ensemble of trained models on a sample from the original training data, we use all of the ensemble residuals to construct a "refinement score" for this sample. This score is then utilized to refine the training data, and isolate the anomalous samples. The USDR framework includes three steps:

\paragraph{Step I: Training a residual-based model on subsets of the training data.} The unlabeled training data $\mathcal{D}_0$ is split into $M$ equally sized \emph{partially overlapping} subsets $\{\mathcal{D}_j\}_{j=1}^M$, such that each sample ${\mathbf x}_i$ appears in $M_{\rm\scriptscriptstyle train}$ of the $M$ sets. The values of $M$ and $M_{\rm\scriptscriptstyle train}$ are determined such that each subset is large enough to allow for training in the known (clean) case, and that both $M_{\rm\scriptscriptstyle train}$ and $M-M_{\rm\scriptscriptstyle train}$ are large enough to allow for statistics, as explained below. Moreover, when defining the subsets, we ensure that each sample is represented in an equal number of training subsets. 

In the present paper we focus on detecting contextual anomalies in time series data. In this case, a simple way to construct the training subsets $\{\mathcal{D}_j\}_{j=1}^M$ is by sliding a window of $w$ samples with a stride of $d$ on the entire time ordered training data. In order to make sure that every sample is equally represented in the training subsets, we use periodic boundary conditions on the training data $\mathcal{D}_0$ when constructing the overlapping subsets. The parameter $d$ is chosen to guarantee large enough subset numbers $M_{\rm\scriptscriptstyle train}$ and $M-M_{\rm\scriptscriptstyle train}$.

After obtaining the subsets $\{\mathcal{D}_j\}$, we train a selected  model $f$ on the inputs ${\mathbf x}_i \in \mathcal{D}_j$ for a given $j$, $1 \le j\le M$ to predict the targets ${\mathbf y}_i$. The model $f$ is expected to act as a residual-based AD model when trained with normal data. However, here we train it with unlabeled data, which may contain abnormal samples. We denote with $f_j$ the model trained with subset $\mathcal{D}_j$. After training on all subsets, we end up having an ensemble of trained models $\{f_j\}_{j=1}^M$, each of which was trained with a subset of the original unlabeled training data. We note again that a given sample 
${\mathbf x}_i$ is included only in $M_{\rm\scriptscriptstyle train}$ of the $M$ subsets, and not in all of them (see Figure \ref{fig:framework}).

\paragraph{Step II: Using the trained ensemble for inference on the entire data.}
In this step we use the trained ensemble $\{f_j\}_{j=1}^M$ to infer on the entire unlabeled dataset $\mathcal{D}_0$. We denote the prediction of model $f_j$ using the input ${\mathbf x}_i$ by $\hat{\mathbf y}_{ij}$:
\begin{equation}
    \hat{\mathbf y}_{ij}=f_j({\mathbf x}_i).
\end{equation}

\paragraph{Step III: Assigning a refinement score to each sample.}
In this step we use the predictions $\hat{\mathbf y}_{ij}$ to assign a refinement score $S^{\rm\scriptscriptstyle USDR}_i$ for each sample ${\mathbf x}_i \in \mathcal{D}_0$. To this end, we first separately rescale the residuals of each individual member $j$ of the ensemble. The rescaled residual of sample $i$ with model $j$ is defined as: 
\begin{equation}
    r_{ij} \equiv \frac{\lvert{\mathbf y}_i-\hat{\mathbf y}_{ij}\lvert-\mu_j}{\sigma_j}
\end{equation}
where
\begin{align}    
    \mu_j &= \frac{1}{N_{j}}\sum_{{\mathbf x}_i\in \mathcal{D}_j} \lvert{\mathbf y}_i-\hat{\mathbf y}_{ij}\lvert\\
    \sigma_j^2 &= {\rm VAR} \left(\lvert {\mathbf y}_i-\hat{\mathbf y}_{ij}\rvert \right) %\frac{1}{N_{j}}\sum_{{\mathbf x}_i\in \mathcal{D}_j} ({\mathbf y}_i-\hat{\mathbf y}_{ij}-\mu_j)^2
\end{align}
are the mean and variance of the residuals within the training subset $\mathcal{D}_j$.

Next, each sample in the original training data obtains a refinement score based on its individual contribution to the training generalization ability. To this end, we define two types of residual means for each sample of the training dataset $\mathcal{D}_0$. The mean residual of sample ${\mathbf x}_i$ over all the models $f_j$ trained with a subset  $\mathcal{D}_j$ which \textit{includes} the sample ${\mathbf x}_i$ is defined as:

\begin{equation}
    z_i^{\rm\scriptscriptstyle (train)}\equiv \frac{1}{M_{\rm\scriptscriptstyle train}}\sum_{j=1}^{M_{\rm\scriptscriptstyle train}}r_{ij}, {\mathbf x}_i\in \mathcal{D}_j
\end{equation}

The mean residual of sample ${\mathbf x}_i$ over all the models $f_j$ trained with a subset  $\mathcal{D}_j$ which \textit{does not include} the data point ${\mathbf x}_i$ (i.e ${\mathbf x}_i$ is outside the training subset) is defined as:
\begin{equation}
    z_i^{\rm\scriptscriptstyle (infer)}\equiv \frac{1}{M-M_{\rm\scriptscriptstyle train}}\sum_{j=1}^{M-M_{\rm\scriptscriptstyle train}}r_{ij}, {\mathbf x}_i\notin \mathcal{D}_j
\end{equation}

The anomaly score of the sample ${\mathbf x}_i$ is then defined as the difference between the two means: 

\begin{equation}\label{eqn:refinement_score}
   S^{\rm\scriptscriptstyle USDR}_i \equiv z_i^{\rm\scriptscriptstyle (infer)}-z_i^{\rm\scriptscriptstyle (train)}
\end{equation}
In other words, the anomaly score of a sample ${\mathbf x}_i$ quantifies the generalization ability of a model which includes ${\mathbf x}_i$ in its training set. If an input sample ${\mathbf x}_i$ is an anomaly, learning its characteristics is likely to add valuable information to the model. We thus expect a significantly lower residual when inferring on ${\mathbf x}_i$ if this sample is in the training set than if it is not included in the training set. On the other hand, for a normal sample ${\mathbf x}_i$, belonging to the dominant normal class, we expect no significant difference, whether the model is trained on this very input or not, due to its similarity to a large number of other samples in the full dataset $\mathcal{D}_0$. We quantify this idea by scoring each sample by its relative contribution to the generalization power of the model. In this way, our anomaly score takes into account not only the properties of the dominant normal class but exploits information from the abnormal samples as well.

\subsection{Baselines}
To evaluate the performance of the proposed USDR framework we compare the results with two generic approaches. 

%\paragraph{}
 
\paragraph{Blind Training.}

 As a naive baseline we use the standard approach in the absence of any data refinement method. We train the residual-based model $f$ "blindly" on the entire training data, despite the anomaly contamination.  We then use the "blindly" trained model to infer on the same data, and obtain scores that should be used to separate normal from abnormal samples. The anomaly score for sample ${\mathbf x}_i$ is its absolute prediction residual:
    \begin{equation}\label{eqn:residuals_blind}
        r_i=\lvert{\mathbf y}_i-  \hat{{\mathbf{y}}}_i\lvert
    \end{equation}
    Since the model $f$ should typically be trained on normal data only, its AD performance is expected to be poor when it is trained on contaminated data.
    The comparison with blind training serves to assess the improvement obtained by the refinement step we propose. In order to isolate the effect of our suggested refinement method from potential effects of the time-ordered subset generation process, we perform the blind training procedure using the same subset split as the USDR training. The anomaly score of each sample is the ensemble mean of the normalized prediction residuals of all models trained with the respective training subsets: 
    \begin{equation}\label{eqn:score_blind_ensemble}
    z_i =  \frac{1}{M}\sum_{j=1}^{M}r_{ij}
\end{equation}
For the sake of method comparability, the scores are then rescaled between 0 and 1.
%%%%%%%%
\begin{figure}[tb]
\begin{center}
\includegraphics[width=\columnwidth]{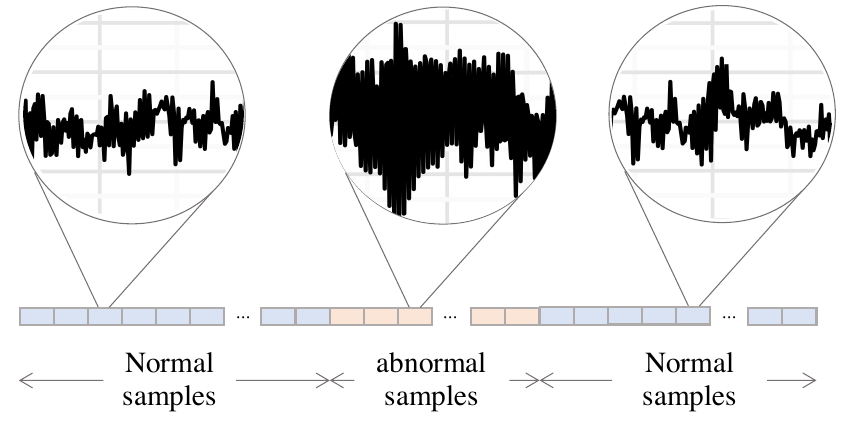}
\caption{MIMII experiment design.}
\label{fig:use_case_MIMII}
\end{center}
\end{figure}
%%%%%%%%%

%\paragraph{}
%
\begin{figure*}[h]
\centering
\includegraphics[width=0.75\textwidth]{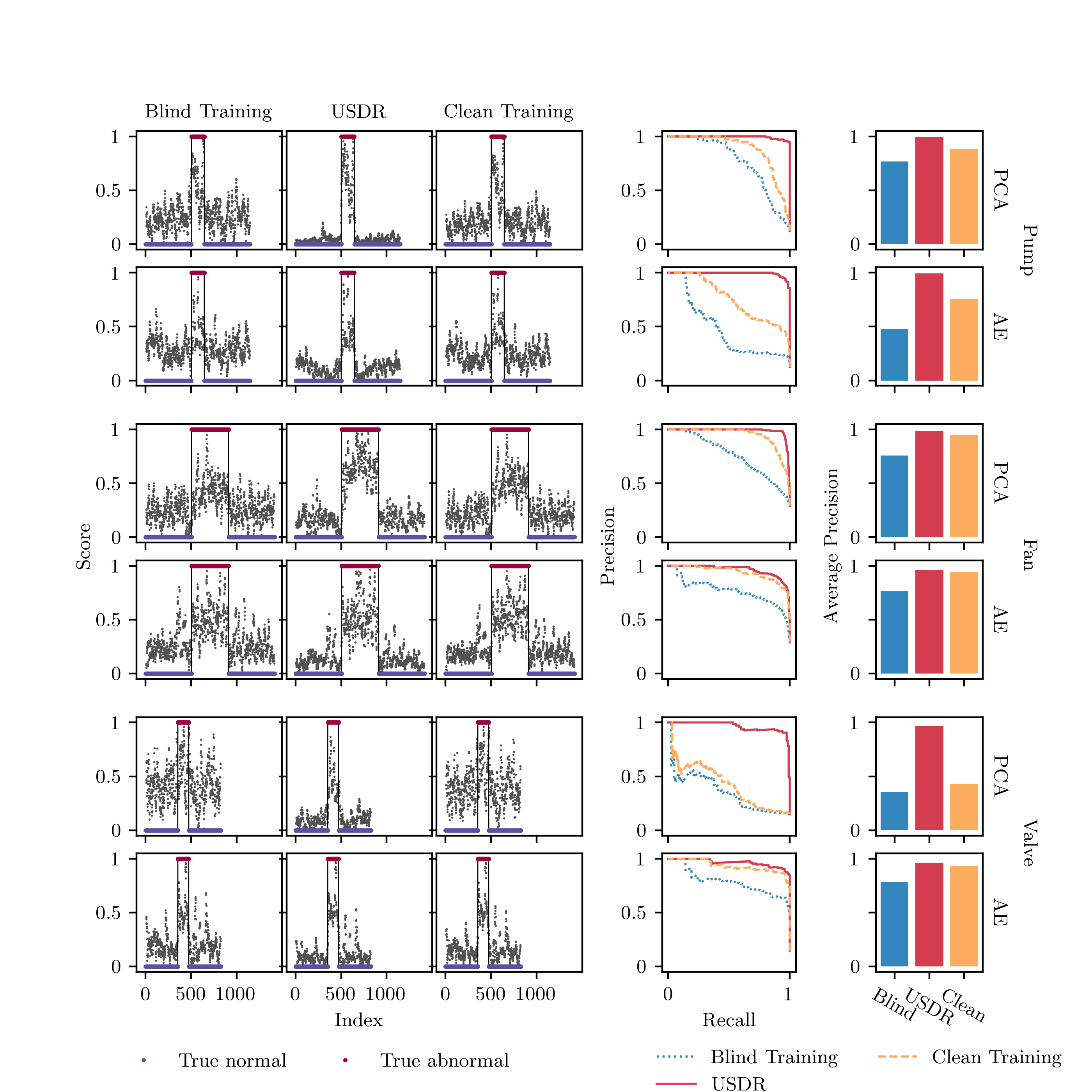}
\caption{Examples of the framework performance for the MIMII data (test case I). The derived scores of the USDR framework are compared with the scores of blind training with the contaminated data and to clean training with normal data as a reference, calculated using PCA (upper row) and AE (lower row). The results are shown for selected cases: Pump (id00,0dB), fan (id00,6dB), and valve (id02,0dB). The two columns on the right show the precision-recall curves (PRC) and the average precision (AP) for the three methods. }
\label{fig:single_fault}
\end{figure*}
%%%%

\paragraph{Clean Training.} 
In an ideal but rather unrealistic case, the training dataset would be manually cleaned prior to training, such that it contains only normal samples. In order to obtain a reference for the difficulty of the refinement task (i.e how easy the separation between normal and abnormal samples is), it is useful to train the selected prediction model $f$ on such an optimally cleaned data (as an assumed best case scenario), and use the trained model to infer on the entire training set $\mathcal{D}_0$. To isolate the effect of the proposed method from potential effects of the subset generation, The anomaly scores are derived in the same way as for the blind model, using Eqn. \ref{eqn:score_blind_ensemble}, however with the healthy training subsets only.

The comparison of the USDR performance with the two extreme baselines is done at the level of the refinement efficacy of all three approaches, by comparing the derived scores on the original training dataset. In this way we explicitly avoid selecting a refinement threshold, a task which would strongly depend on the selected model. The latter will be addressed in a separate study, focusing on state-of-the-art AD performance tests.

\begin{figure*}[h]
\begin{center}
\includegraphics[width=0.75\textwidth]{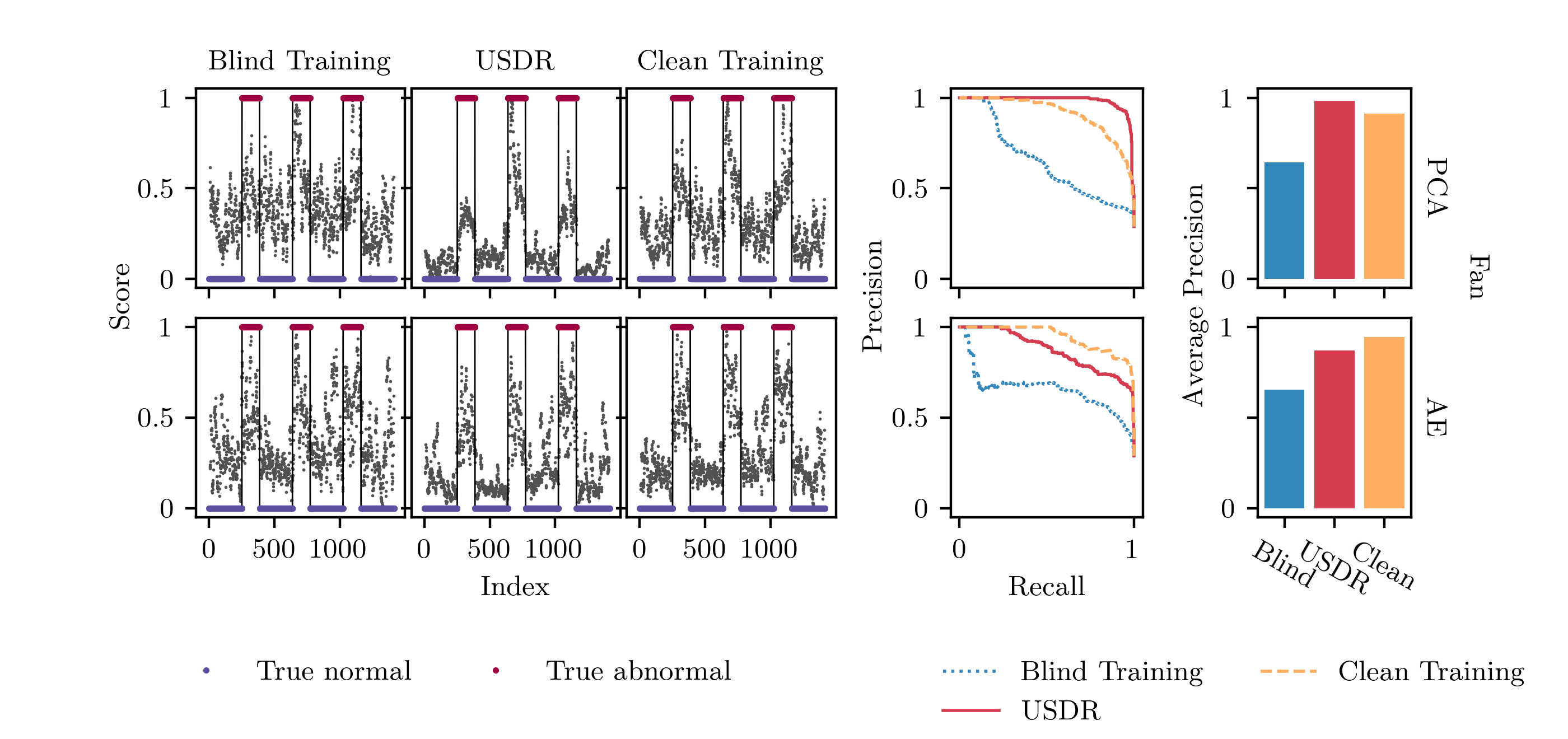}
\caption{Examples of the framework performance for the MIMII data (test case II). The figure structure is similar to Fig. \ref{fig:single_fault}, demonstrated for the Fan system. Here we assumed three short faulty periods instead of a single long fault.}
\label{fig:multiple_faults}
\end{center}
\end{figure*}

%%%%%%%%%%%%%%%%%%%%%%%%%%%%%%%%%%%%%
%%%%%%%%%%%%%%%%%%%%%%%%%%%%%%%%%%%%%
\section{Experiments}\label{sec:results}
%%%%%%%%%%%%%%%%%%%%%%%%%%%%%%%%%%%%%
%%%%%%%%%%%%%%%%%%%%%%%%%%%%%%%%%%%%%
In order to demonstrate the performance of our suggested method, we conduct tests on two public datasets of technical machine data. The first dataset is the sound data for Malfunctioning Industrial Machine Investigation and Inspection (MIMII) \cite{purohit2019mimii}, containing acoustic signals of normal and abnormal machine components. With this data we construct a use case that mimics the abrupt appearance of faults in a previously healthy machine. The second dataset, the Turbofan engines N-CMAPSS DS02 of NASA \cite{saxena2008} contains full degradation trajectories of aircraft engines. The two examples differ not only in the data type and physical context but also in their different fault dynamics, which is abrupt in one case vs. slowly degrading in the other. 

The purpose of the evaluation below is to demonstrate the generic nature of our proposed approach, which does not depend on the dataset nor on the selected model $f$. The only prerequisite is that $f$ is a residual-based model. We show that the unsupervised data refinement (USDR) step significantly improves the performance of the model $f$ compared to the alternative of training blindly on contaminated data ("blind training"). As a reference for the separability of normal and abnormal samples, we show in each case the results of training with the anomaly-free part of the data ("clean training"). 

It is important to note that the goal of this work is not to obtain state-of-the-art AD performance on the analyzed datasets, but rather to demonstrate the benefit achieved by applying the refinement framework, independent of the chosen prediction model and of the dataset. Therefore, we do not perform extensive studies to select the best performing model, but rather demonstrate the performance on two of the most popular AD models, PCA and fully connected Autoencoder (AE). The model parameters are minimally adjusted to achieve good performance for the ideal "Clean Training" case. Since the framework is model agnostic, we expect a similar performance enhancement due to the data refinement step, if one replaces these basic standard models with state of the art alternatives (e.g more complex, including generative AE variants), depending on the use case at hand.

%%%%%%%%%%%%%%%%%%%%%%%%%%%%%%%%%%%%%%%%%%
\subsection{Acoustic Signal Data}
\label{subsec:mimii}
%%%%%%%%%%%%%%%%%%%%%%%%%%%%%%%%%%%%%%%%%%

\paragraph{Description of the dataset.}

The MIMII dataset \cite{purohit2019mimii} contains audio recordings from four types of machines: fans, pumps, valves and slide rails. Some of the recordings were taken when the machines were normally functioning and others in malfunctioning states. For each machine type, normal and abnormal data from 4 individual units is recorded, with no further labels of the fault type of operating condition. For each unit, data under 3 signal-to-noise (SNR) levels (6, 0 and -6dB), controlling the background noise (unrelated to the machine functioning state). Similar to the original paper \cite{purohit2019mimii}, we use only the first channel of microphones, and consider log-Mel spectrograms as the input feature to the AD models. The spectrograms are generated as described in the original dataset description \cite{purohit2019mimii}.

\paragraph{Experiment design.}

We use the MIMII dataset to mimic a realistic industrial setting in which continuously recorded condition monitoring data from a given machine is contaminated with unlabeled faults. To this end, we concatenate all of the acoustic signals from the MIMII dataset which belong to a single individual unit (under fixed SNR conditions). We would like to approximate a situation in which a machine is turning faulty (malfunctioning) in an abrupt way, and is then being repaired thus regaining its normal functioning. This results in a time series that contains a first normal (healthy) period (constructed by concatenating $n_0$ acoustic signals), then one (or more) faulty periods followed by normal periods due to the repair. Each normal (abnormal) period is constructed by signal concatenation of $n_h$ ($n_f$) acoustic signals. An example of how the time series of concatenated signals is constructed is shown in Figure \ref{fig:use_case_MIMII}. We note that the faulty period may contain a mixture of failure modes, as the original dataset does not contain labels of the fault types or root causes. 

\paragraph{Test cases.}

We consider two test cases, mimicking two different anomaly structures in the MIMII training data. In test case I, we construct a condition monitoring signal with a single faulty period, and in test case II there are 3 short faulty periods containing abnormal signals. The purpose of this is to demonstrate the effectiveness of the framework for different contextual anomalies, approaching the limit of point anomalies. The latter will be further tested in a separate study. 

Within each test case we use the USDR framework to refine the entire data available in the dataset for a specific unit. We repeat this for all units of the 4 machine types: fan, pump, slide rail, and valve.

To this end, the available data for a given unit is split using a sliding window of a fixed length of $200$ samples into  partially overlapping training sets, such that each sample repeats in $M_{\rm\scriptscriptstyle train}=5$ sets.

\paragraph{Results.}

\begin{table*}[h]
    \centering
    \ra{1.3}
    \caption{Average precision (AP) scores for AD on contaminated MIMII data.}
    \label{table:MIMII_results}
    \begin{tabular}{@{}lcllllclllcccc@{}}
        \toprule
        &&& \multicolumn{3}{c}{PCA} & \phantom{X}& \multicolumn{3}{c}{AE} &
        \phantom{X} & \multicolumn{1}{c}{OC-SVM}& \phantom{X} &\multicolumn{1}{c}{IF}\\
        \cmidrule(lr){4-6} \cmidrule(lr){8-10} \cmidrule(lr){12-12}\cmidrule(lr){14-14}
        &Contamination&& Blind  & USDR & Clean  && Blind & USDR & Clean &&    &&   \\ 
        \midrule
        %Fan\\
        &&6 dB & $0.94$ & $1.00$ & $0.99$ && $0.90$ & $0.98$ & $0.97$ && $0.86$ && $0.97$  \\
        Fan& 29\% &0 dB & $0.75$ & $0.93$ & $0.85$ && $0.73$ & $0.83$ & $0.83$ && $0.60$ && $0.77$  \\
        &&-6 dB & $0.55$ & $0.67$ & $0.60$ && $0.50$ & $0.63$ & $0.64$ && $0.37$ && $0.36$  \\
        \midrule
        %Pump\\
        &&6 dB & $0.95$ & $1.00$ & $0.98$ && $0.96$ & $1.00$ & $0.99$ && $0.61$ && $0.90$  \\
        Pump& 12\% &0 dB & $0.71$ & $0.99$ & $0.84$ && $0.65$ & $0.91$ & $0.83$ && $0.5$ && $0.67$  \\
        &&-6 dB & $0.39$ & $0.77$ & $0.45$ && $0.41$ & $0.58$ & $0.52$ && $0.22$ && $0.41$  \\
        \midrule
        %Slider\\
        &&6 dB & $0.77$ & $0.98$ & $0.95$ && $0.82$ & $0.91$ & $1.00$ && $0.46$ && $0.80$  \\
        Slider& 25\% &0 dB & $0.60$ & $0.94$ & $0.77$ && $0.88$ & $0.96$ & $0.99$ && $0.38$ && $0.75$  \\
        &&-6 dB & $0.35$ & $0.60$ & $0.50$ && $0.72$ & $0.80$ & $0.87$ && $0.23$ && $0.32$  \\
        \midrule
        %Valve\\
        &&6 dB & $0.41$ & $0.66$ & $0.49$ && $0.79$ & $0.92$ & $0.90$ && $0.31$ && $0.46$  \\
        Valve& 11\%&0 dB & $0.20$ & $0.47$ & $0.24$ && $0.56$ & $0.70$ & $0.71$ && $0.24$ && $0.31$  \\
        &&-6 dB & $0.12$ & $0.14$ & $0.12$ && $0.27$ & $0.48$ & $0.43$ && $0.13$ && $0.09$  \\
        \bottomrule
    \end{tabular}
\end{table*}

Selected results of test case I are shown in Figure \ref{fig:single_fault}. The three columns on the left show the derived scores for the three methods: blind training, USDR (the proposed framework), and clean training. The scores are shown for three different units of three machine types (fan, pump, and valve) with two simple reconstruction models, PCA and fully connected AE, for each machine type. In all cases we trained a PCA with 5 principal components and an AE with 7 dense layers, a latent dimension of 80 (resulting in 236944 trainable parameters) and ReLU activations. In each panel, the true labels are marked in color for reference: blue for normal samples with label $0$ and red for abnormal ones with label $1$. 
The displayed scores have a different derivation for each of the three methods. For Blind Training and Clean Training, we show the anomaly scores obtained by rescaling the reconstruction errors (Eqn. \ref{eqn:residuals_blind} with ${\mathbf y}_i={\mathbf x}_i$ and $\hat{{\mathbf{y}}}_i=\hat{{\mathbf{x}}}_i $) to range between $0$ and $1$. For the USDR framework we show the refinement scores $S^{\rm\scriptscriptstyle USDR}_i$ obtained using Eqn. \ref{eqn:refinement_score}. To improve the separability all scores are smoothed with a moving mean of $10$ samples. We note that this step can be avoided by further optimizing the choice of the prediction model $f$ to ideally fit the use case at hand, a step that we did not undertake here in order to remain simple and emphasize the model independence of our framework. 

For each machine unit and SNR condition, we use all available normal and abnormal samples as the initial training data with the aim of data refinement, i.e. selecting only the normal samples in order to use them to retrain an AD model in the next step. Note that we do not show the results of such a second step. In this way the results we show remain generic, with no need to determine a refinement threshold. In contrast to our proposed framework, the optimal threshold would be model-dependent.  

In addition to showing the derived refinement scores, Figure \ref{fig:single_fault} shows the corresponding precision-recall curves (prc) and their resulting average precision (AP) on the two right-most columns. The prc curves result from evaluating the scores obtained by each of the three methods (blind, USDR, and clean) against the true normal/abnormal labels of the training samples. In this way, the prc measures the quality of the separation obtained by each method between normal and abnormal samples in the contaminated dataset. The corresponding AP values are displayed as bars in the right-most panel of each row. 

\begin{figure*}[tb]
\begin{center}
\includegraphics[width=0.75
\textwidth]{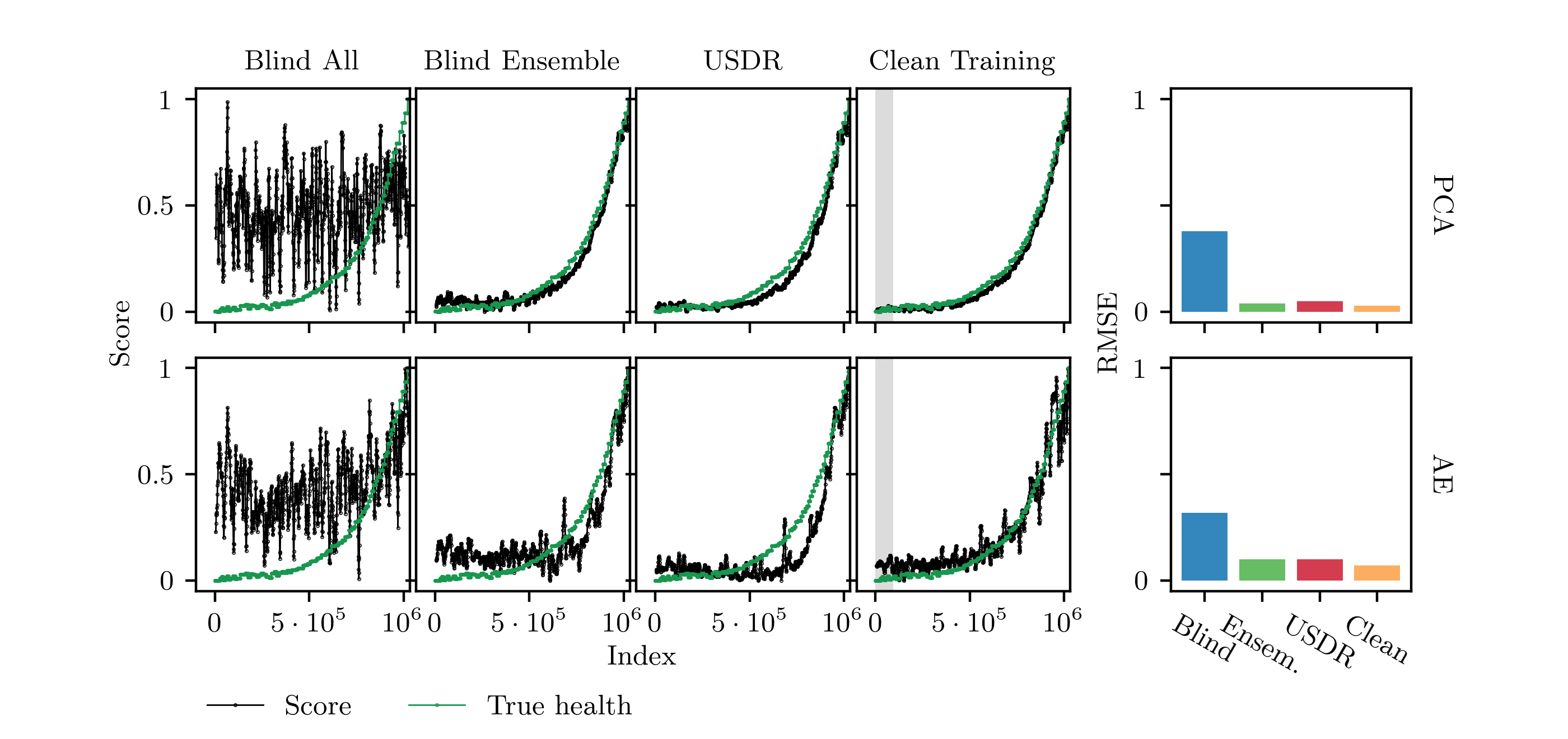}
\caption{Examples for the framework performance with the turbofan engine CMAPSS data. The derived scores of the USDR framework are compared with the scores of blind training with the all contaminated data,  blind ensemble, and clean training with normal data (first 10 engine cycles) as a reference, calculated using PCA (upper row) and AE (lower row). The results are shown for engine 5. The column on the right shows the RMSE for the four methods.}
\label{fig:turbofan_scores}
\end{center}
\end{figure*}

From Figure \ref{fig:single_fault}, it is evident that the Blind Training (left most column) performs poorly for all machine types and with both PCA and AE models. In many cases, it is close to impossible to find a threshold that would lead to a proper refinement of the training data based on the blind anomaly scores. 

The suggested USDR framework (second column from the left), however, shows a clear separation between the refinement scores of true normal and true abnormal samples. This allows to select a threshold that removes most of the abnormal samples without removing many normal samples from the training data, thereby refining the data for subsequent training of a residual-based AD model. The separability obtained by the USDR is comparable to the Clean Training reference (third column from the left), and is often higher. The comparison of the three methods can be quantified in terms of the prc results on the two columns on the right. The USDR method (red) always significantly outperforms the blind training (blue) in the achieved AP, and performs at least as well as the ideal reference trained with anomaly-free data (yellow). In several cases it is seen to outperform the clean reference, which may be interpreted as the "best-case-scenario". This result can be intuitively understood when considering that the USDR method explicitly contrasts the information from abnormal samples with the information from normal ones, by quantifying the contribution of each sample to the generalization ability of the trained model at inference time.  

We note that depending on the machine type, either PCA or AE show better refinement performance. 
A similar model dependence is also observed for the Clean Training anomaly scores displayed in the third column for reference. For example: even in the ideal case of anomaly-free training data, PCA performs quite poorly on the valve data, whereas AE displays a rather high separability between normal and abnormal samples. It is interesting to note, that in the PCA case the benefit of the USDR is particularly high. In this case, the residuals themselves contain little information about the anomaly. However, the drop in the residual of a specific sample once it is included in the training set is the discriminating factor of anomalies from normal samples.

The results of this test case are summarized in Table \ref{table:MIMII_results}. For each machine type and SNR, the mean score over ten repetitions and the four units is shown. This is repeated for the three methods (blind training, USDR and clean training) using both PCA and AE. As a reference for the task complexity, we display the mean score achieved by two simple unsupervised baselines: one-class support vector machine (OC-SVM) and isolation forest (IF). The performance of these two naive approaches deteriorates as the task of separation between the normal and abnormal samples gets more difficult. We note that the OC-SVM baseline was calculated assuming that the anomaly fraction is known (as opposed to the proposed USDR which does not assume this). The table demonstrates a broad range of difficulties of the AD tasks, with various contamination ratios and different fault features for each machine type and SNR level. Accordingly, the relative performance of different AD models can strongly vary. However, the common results to almost all cases is a significant improvement achieved by the USDR method compared to the blind training when both are using the same underlying model (either PCA or AE). The refinement task becomes harder when the SNR ratio decreases. Particularly noisy cases (e.g. Valve with -6dB) are hard to refine, also for the clean training reference model. Another common outcome is the fact that the USDR is on par with the ideal Clean Training case (which would not be available in many real-world applications), and sometimes outperforms it, given a selected model $f$ (here PCA or AE).

We note again, that the purpose of the present study is not to obtain state-of-the-art AD performance, but rather to demonstrate the striking improvement achieved by the suggested refinement framework, compared to the alternative of training with contaminated data. This improvement is shown here across machine types, contamination ratios, and prediction models. The absolute refinement performance can be further improved by optimizing the prediction model at hand, which is beyond the scope of the present work.

Exemplary results from Test case II are displayed in Figure \ref{fig:multiple_faults}. The structure of the figure is identical to the one of Figure \ref{fig:single_fault} and the results are shown for the Fan, containing $29\%$ contamination in the training data. The two cases differ in the way the training samples were concatenated to construct the condition monitoring time series. In test case II we mimic a situation of three short faulty periods instead of a single longer fault. The purpose of this additional test is to demonstrate the performance improvement achieved by the USDR framework, for several different abrupt fault scenarios. The next step in this direction, of addressing the case of completely isolated outliers, is part of our future research.  

Figure \ref{fig:multiple_faults} demonstrates again that the proposed USDR method significantly outperforms the blind training alternative, and often reaches the performance of the unrealistic Clean Training scenario.

\subsection{Aircraft Engine Data}
\label{subsec:turbofan}
%%%%%%%%%%%%%%%%%%%%%%%%%%%%%%%%%%%%%%%%%%
\paragraph{Description of the dataset.}

As a second use case, we evaluate the performance of the proposed framework on the well-known turbofan engine dataset. This dataset contains synthetic run-to-failure degradation trajectories of nine aircraft engines, generated with the Commercial Modular Aero Propulsion System Simulation (CMAPSS) model of NASA \cite{frederick2007user}, and was generated taking real flight conditions from commercial jets as input. The dataset includes 19 variables of the flight conditions as well as the temperatures and pressure levels at various parts of the engine, for multiple flight cycles, from the beginning of life until full degradation of the engine. Different engines (units) display different failure modes, some affecting a single component and others affecting multiple components within the engine \cite{chao2022fusing}. 
\begin{table*}[h]%\begin{table}[htbp]
    \centering
    \caption{RMSE of Anomaly Scores for Turbofan Data}
    \label{table:turbofan_results}
    \begin{tabular}{@{}lcccc cccc@{}}
        \toprule
        Unit & \multicolumn{4}{c}{PCA} & \multicolumn{4}{c}{AE} \\
        \cmidrule(lr){2-5} \cmidrule(lr){6-9}
        & Blind all& Ensemble & USDR & Clean & Blind all & Ensemble & USDR & Clean\\
        \midrule
        2 & $0.34$ &0.06& $0.07$ & $0.03$ & $0.29$ &0.15& $0.13$ & $0.07$ \\
        5 & $0.38$ &0.04& $0.04$ & $0.02$ & $0.32$ &0.10& $0.10$ & $0.05$ \\
        10 & $0.33$ &0.04& $0.05$ & $0.03$ & $0.31$ &0.12& $0.11$ & $0.08$ \\
        \bottomrule
    \end{tabular}
\end{table*}%\end{table}
A standard training procedure with the CMAPSS data uses a residual-based model (often a reconstruction model with an AE architecture) with all 19 time series variables as input and output. In the standard approach, the first flight cycles of a given engine are assumed to represent healthy behavior and are thus used to train the reconstruction model. At inference time, the model is expected to yield large reconstruction errors as the engine condition deteriorates, that is with a growing cycle number. Various extensions of this basic architecture were suggested along the years in order to predict the remaining useful life (RUL) of the various unit engines (see for example \cite{li2018remaining,arias2021aircraft}. Here, however, we do not aim to predict the RUL but use this dataset for the purpose of unsupervised detection of abnormal behavior, observed in this case as a slow degradation of the condition of the engine. In this setting, any deviation from the normal behavior is regarded as "anomaly" with an assigned anomaly score, which ideally reflects the degree of degradation and can thus be converted into a health index. At a later stage, which is beyond the scope of the present paper, the health index may be used to predict the RUL.

\paragraph{Results.} 

Figure \ref{fig:turbofan_scores} displays the refinement scores derived using the USDR framework for the entire degradation trajectory of a single engine, in this case unit 5 of the CMAPSS DS02 dataset. 
As in the MIMII use-case, the available data for this unit is split using a sliding window of length $20\%$ of the available data into $M=20$ partially overlapping training subsets, such that each sample repeats in $M_{\rm\scriptscriptstyle train}=4$ subsets. 
Here as well, we contrast the refinement scores of the USDR with the anomaly scores (rescaled reconstruction errors) of the Blind Training approach and the Clean Training for reference. In this case we use "Blind All" to refer to a single training with the entire data set, and "Blind Ensemble" to refer to training an ensemble of models, for which the ensemble mean of the normalized residuals is used as a score. The Clean Training scores are obtained by training the reconstruction models with data from the first 10 engine cycles (which are assumed to be degradation-free and are marked with a grey background) and inferring on the full degradation data. In each case we trained two simple residual-based models, PCA and a fully connected AE. 
For the PCA we select 15 principal components, and the AE is trained with 5 dense layers, a latent dimension of 20 (resulting in 6908 trainable parameters), and ReLU activations.

In all cases we rescale the scores between 0 and 1 and compare them with the true health index provided for this dataset (green). The left column of Figure \ref{fig:turbofan_scores} shows clearly that blindly training a reconstruction model using the full degradation trajectory with no refinement fails to reveal the true health condition of the unit. However, both the Blind Ensemble (second column from the left) and the USDR score (third column from the left) follow the true health index rather accurately, in particular using a simple PCA model, allowing to clearly separate healthy from degraded data. Indeed, as shown in the rightmost column, the RMSE of these two scores with respect to the true health index is almost as small as the one of the Clean Training reference (fourth column from the left). It is worth noting that for this dataset, the USDR and the simpler ensemble refinement are similarly powerful in discovering the health condition of the engine in a fully unsupervised manner. They reach remarkably low RMSEs considering the high data contamination: only around $20\%$ of the entire training data can be considered completely "normal". The success of both ensemble-based approaches to discover the degradation pattern potentially hinges on the similarity between healthy and degraded conditions (as we know that only very subtle degradation effects were injected in the simulation), except for towards the very end of the degradation trajectory.

Table \ref{table:turbofan_results} shows the AD performance for three different engines. In order to compare with a clear ground truth health indicator we selected the three engines with a simple degradation mode, of an abnormal high pressure turbine (HPT) efficiency degradation, for which the true health index can be easily extracted. The Table displays the RMSE of the different scores (Blind All, Blind Ensemble, USDR and clean training) with respect to the true health index, each with both PCA and AE as a reconstruction model (where the mean over 10 repetitions is shown). A similar conclusion follows for all three units: given a reconstruction model (PCA or AE), both refinement frameworks allow to identify the normal and abnormal segments of the data in a fully unsupervised way, achieving an RMSE which is only sightly higher than the ideal reference (anomaly-free training data). The improvement over training blindly with the entire data is striking, reducing the RMSE from around $0.3$ or $0.4$ down to around $0.05$.

%\tbl{This is the caption for this Table in font 9pt.  Short line
%caption must be centered.}%\label{tab1}}
%\caption{Caption}

%%%%%%%%%%%%%%%%%%%%%%%%%%%%%%%%%%%%%
%%%%%%%%%%%%%%%%%%%%%%%%%%%%%%%%%%%%% 
\section{Conclusions}
%%%%%%%%%%%%%%%%%%%%%%%%%%%%%%%%%%%%%
%%%%%%%%%%%%%%%%%%%%%%%%%%%%%%%%%%%%%
In this paper we suggest a novel data centric approach to deal with the challenge of AD without any labels, and with potentially contaminated training data. The problem is highly relevant in real-world scenarios, where labeling is expensive and sometimes impractical, and where mislabeling is a common issue. The proposed USDR framework avoids assuming that the training data is anomaly-free but rather allows for an unknown fraction of anomalies, possibly of various types and severities. It suggests a fully-unsupervised refinement of the training data, based on training any residual-based (reconstruction or regression) model on partially overlapping subsets of the contaminated training set. A refinement score is derived for each training sample based on its contribution to the generalization ability of the model, which is shown to increase significantly for abnormal samples. The advantage of the proposed approach lies in its simplicity and generic data-centric nature; it is model agnostic and is conceptually applicable to any data type. It can be applied either to the raw samples or to their learnable representations. We demonstrate the refinement efficacy of USDR for AD of contextual anomalies in multivariate time-series data from industrial machines (contamination fraction up to $29\%$) and aircraft engines (with only $20\%$ normal data). We show that it performs similarly and sometimes better than a model trained on anomaly-free data. In this paper we focus on demonstrating the generic character of the framework, leaving the comparison with state-of-the-art AD models to future work. Extending the applicability of the proposed method to other data modalities (e.g. image data) and testing its hyperparameter sensitivity are additional topics for future research. 

%%%%%%%%%%%%%%%%%%%%%%%%%%%%%%%%%%%%%
%%%%%%%%%%%%%%%%%%%%%%%%%%%%%%%%%%%%% 
%\section*{Acknowledgment}
%%%%%%%%%%%%%%%%%%%%%%%%%%%%%%%%%%%%%
%%%%%%%%%%%%%%%%%%%%%%%%%%%%%%%%%%%%% 

%\section*{References}
%\bibliographystyle{chicago}
%\bibliography{References}

% Bibliography
% ---------------------------------------------------------------------------------
\bibliographystyle{apacite}
\PHMbibliography{References}
% ---------------------------------------------------------------------------------

\end{document}